\documentclass[reprint,amsmath,amssymb,aps]{revtex4-1}
\usepackage{bm}

\begin{document}

\title{Vulnerability of Deep Learning}

\author{Richard Kenway}
 \email{Email address: r.d.kenway@ed.ac.uk}
\affiliation{School of Physics and Astronomy, University of Edinburgh, James Clerk Maxwell Building, Peter Guthrie Tait Road, Edinburgh EH9 3FD, UK}

\date{\today}

\begin{abstract}
The Renormalisation Group (RG) provides a framework in which it is possible to assess whether a deep-learning network is sensitive to small changes in the input data and hence prone to error, or susceptible to adversarial attack. Distinct classification outputs are associated with different RG fixed points and sensitivity to small changes in the input data is due to the presence of relevant operators at a fixed point. A numerical scheme, based on Monte Carlo RG ideas, is proposed for identifying the existence of relevant operators and the corresponding directions of greatest sensitivity in the input data. Thus, a trained deep-learning network may be tested for its robustness and, if it is vulnerable to attack, dangerous perturbations of the input data identified.  
\end{abstract}


\maketitle

\section{\label{sec:introduction} Introduction}

\subsection{\label{subsec:threat} The Threat to Deep Learning}

Despite many successful applications, deep learning~\cite{HintonOsinderoTeh2006, LeCunBengioHinton2015, Salakhutdinov2015} is poorly understood. The absence of an underlying theory means that we cannot be certain under what circumstances a given trained network will operate correctly and this uncertainty calls into question the use of such networks in safety-critical applications. There is also the possibility that slightly perturbed input data can be constructed that would fool the network, but not a human, making the network susceptible to adversarial attack, or simply cause it to misclassify and thereby conceal information in the data~\cite{YuanHeZhuBhatLi2018}.

It is important to distinguish between ambiguous data, which any system, machine or human, would find difficult to classify and data that has been imperceptibly altered to cause the deep network to misclassify it. This paper introduces a theoretical framework, based on the Renormalisation Group (RG), which specifies the conditions under which the latter may occur and proposes a calculational method for testing a trained network for this vulnerability. Further, it determines the directions in the space of input data along which small perturbations are amplified by such a vulnerable deep network, causing it to generate a qualitatively incorrect output.

The theoretical framework captures the salient features of deep learning: the multi-layer structure, trained one layer at a time. It is demonstrated for the specific case in which each layer is a Restricted Boltzmann Machine (RBM)~\cite{Salakhutdinov2015} and the network is trained to classify a data set $\{\bf x\}$ in terms of outputs $\{\bf y\}$, whose exact conditional probability distribution is $p(\bf y|\bf x)$. However, the approach may be extended straightforwardly to other deep networks and to data generation as well as classification.

\subsection{\label{subsec:RG} Analogy with the Renormalisation Group}

The formal analogy between deep learning and the RG has been noted before, although with some controversy about whether the context should be unsupervised, or supervised learning~\cite{MehtaSchwab2014, LinTegmarkRolnick2017, SchwabMehta2016}. There are more fundamental issues.

The first is that the RG applied to critical phenomena describes how features of the microscopic degrees of freedom of a system (analogous to the input data $\{\bf x\}$) depend on length scale, successively eliminating short-distance fluctuations and generating a sequence of effective Hamiltonians, which describe the surviving features on larger and larger length scales. A critical point is associated with scale-invariant fluctuations and convergence of the sequence of effective Hamiltonians to a fixed-point Hamiltonian which describes them. Although many applications of deep learning do involve data which exhibit structure on different length scales, e.g., data drawn from the physical world, and this appears to be captured by deep learning so that it is ``cheap''~\cite{LinTegmarkRolnick2017}, this need not be the case.

The second is that the RG transformation effected by each layer of the network may be different, reflecting the emergence of different types of features in the data, whereas in critical phenomena each application of the RG transformation changes the length scale by a fixed amount. 

We will assume that training of the deep network converges to a conditional probability distribution that is a good approximation to $p(\bf y|\bf x)$, provided there is sufficient number of layers. Thus, the sequence of RG transformations corresponding to applying the RBM in each layer to the output from the hidden nodes in the previous layer, starting with the input data, converges. 

Presumably, some more generalised form of scaling takes place in deep learning and we will assume that this has the effect of successively scaling to smaller values all but a finite number of eigenvalues of the Fisher Information Matrix (FIM) of the trained deep network, so that its spectrum is sloppy~\cite{MachtaChachraTranstrumSethna2013}. We will say more about this in Section~\ref{sec:data}.

In Section~\ref{sec:RGforDL} we introduce the RG formalism in the context of a deep RBM network used for classification and, in particular, define the sequence of effective Hamiltonians whose fixed points and universality classes define distinct outputs. In Section~\ref{sec:exponents} we develop the computational scheme, based on Monte Carlo RG~\cite{Ma1976, Swendsen1979}, for estimating the stability of these fixed points with respect to the depth of the network. Section~\ref{sec:data} relates the instability of a particular fixed point to direction(s) in the input data space $\{\bf x\}$, along which small changes in the data cause major changes in the output probability distribution and, hence, are sources of vulnerability.

\section{\label{sec:RGforDL} Renormalisation Group for Deep Learning}

Consider a layered network of $N$ RBMs with input nodes ${\bf h}_0$, $N-1$ layers of hidden nodes ${\bf h}_k$, $k=1,\ldots,N-1$, and output nodes ${\bf h}_N$. For example, the hidden nodes may be binary vectors of the same dimension as the input data $\{\bf x\}$ and the outputs $\{\bf y\}$, and the joint probability distribution for the $k^{\rm th}$ layer is of the form~\cite{Salakhutdinov2015}
\begin{eqnarray}
t_k({\bf h}_k, {\bf h}_{k-1}) &=& \frac{{\rm e}^{{\bf h}_k^{\rm T} {\bf W}_k {\bf h}_{k-1} + {\bf a}_k^{\rm T}{\bf h}_k + {\bf b}_k^{\rm T}{\bf h}_{k-1}}}{z_k},\nonumber\\ 
z_k &=& {\rm Tr}_{{\bf h}_k, {\bf h}_{k-1}}{\rm e}^{{\bf h}_k^{\rm T} {\bf W}_k {\bf h}_{k-1} + {\bf a}_k^{\rm T}{\bf h}_k + {\bf b}_k^{\rm T}{\bf h}_{k-1}},
\label{eq:RBM}
\end{eqnarray}
where ${\bf W}_k$, ${\bf a}_k$ and ${\bf b}_k$ are weights determined by the layerwise training. We will not require the precise form of $t_k$.

The trained deep RBM network generates the probability distribution for the output $\bf y$ given input data $\bf x$, $q_N({\bf y}|{\bf x})$, iteratively for $k=1,\ldots,N$ through
\begin{eqnarray}
q_k({\bf h}_k|{\bf x}) &=& {\rm Tr}_{{\bf h}_{k-1}}t_k({\bf h}_k|{\bf h}_{k-1})q_{k-1}({\bf h}_{k-1}|{\bf x}),\nonumber\\
q_0({\bf h}_0|{\bf x}) &=& \delta_{{\bf h}_0, {\bf x}},\nonumber\\
q_N({\bf y}|{\bf x}) &\approx& p({\bf y}|{\bf x}),
\label{eq:deepRBMprob}
\end{eqnarray}
where, according to Bayes Theorem,
\begin{equation}
t_k({\bf h}_k|{\bf h}_{k-1}) = \frac{t_k({\bf h}_k, {\bf h}_{k-1})}{{\rm Tr}_{{\bf h}_{k}}t_k({\bf h}_k,{\bf h}_{k-1})}.
\end{equation}

We define the effective Hamiltonian for the $k^{\rm th}$ layer by
\begin{eqnarray}
q_k({\bf h}|{\bf x}) &=& \frac{{\rm e}^{-H^{(k)}_{\bf x}({\bf h})}}{Z^{(k)}_{\bf x}},\nonumber\\
Z^{(k)}_{\bf x} &=& {\rm Tr}_{\bf h} {\rm e}^{-H^{(k)}_{\bf x}({\bf h})}.
\label{eq:Hamiltonian}
\end{eqnarray}
Then, the sequence of Hamiltonians is generated by the RBMs in each layer via
\begin{equation}
\frac{{\rm e}^{-H^{(k+1)}_{\bf x}({\bf h})}}{Z^{(k+1)}_{\bf x}} = {\rm Tr}_{{\bf h}'}t_{k+1}({\bf h}|{\bf h}')\frac{{\rm e}^{-H^{(k)}_{\bf x}({\bf h}')}}{Z^{(k)}_{\bf x}},
\label{eq:RGT}
\end{equation}
such that each Hamiltonian depends only on the previous one in the sequence. This has the same form as an RG transformation with kernel $t_{k+1}$~\cite{MehtaSchwab2014}.

At this point, it is worth noting that the input data $\bf x$ determines the couplings in the space of effective Hamiltonians via the first layer,
\begin{equation}
\frac{{\rm e}^{-H^{(1)}_{\bf x}({\bf h})}}{Z^{(1)}_{\bf x}} = t_1({\bf h}|{\bf x}),
\label{eq:first layer}
\end{equation}
and the couplings in each subsequent layer are functions solely of the couplings in the previous layer.

We parametrise the space of effective Hamiltonians in terms of a complete set of operators $O_\alpha({\bf h})$, which, in the case of binary variables, consist of all possible products of the components of ${\bf h}$, with couplings $g_\alpha$. Thus, the effective Hamiltonian for layer $k$ is
\begin{equation}
H^{(k)}_{\bf x}({\bf h}) = \sum_\alpha{g^{(k)}_\alpha O_\alpha({\bf h})}.
\label{eq:couplings}
\end{equation}
Then the effect of the RG transformation in eq.~(\ref{eq:RGT}) is to define $\{g^{(k+1)}\}$ solely in terms of $\{g^{(k)}\}$ and, thereby, to generate a flow in the coupling-constant space of the effective Hamiltonians.

The key assumption we make is that the training of the deep RBM network converges, so that, for $N$ large enough, the sequence of effective Hamiltonians converges to a fixed point: 
\begin{equation}
H^*_{\bf x}({\bf h}) = \sum_\alpha{g^*_\alpha O_\alpha({\bf h})},
\label{eq:fixed point Hamiltonian}
\end{equation}
such that
\begin{equation}
\frac{{\rm e}^{-H^*_{\bf x}({\bf y})}}{Z^*_{\bf x}} = q_N({\bf y}|{\bf x}) \approx p({\bf y}|{\bf x}).
\label{eq:fixed point}
\end{equation}

The analogue of the RG universality class of the fixed point $H^*_{\bf x}$ is the subset of the input data, $C_{\bf x}$, which are associated with the same output distribution, $p({\bf y}|{\bf x})$. Thus, assuming the training correctly classifies the data,
\begin{equation}
{\bf x}, {\bf x}' \in C_{\bf x} \Rightarrow H^*_{\bf x} = H^*_{{\bf x}'}
\label{eq:universality class}
\end{equation}
and
\begin{equation}
p({\bf y}|{\bf x}) \neq p({\bf y}|{\bf x}') \Rightarrow H^*_{\bf x} \neq H^*_{{\bf x}'},
\label{eq:different fixed points}
\end{equation}
so that there must be more than one fixed point. In any non-trivial classification problem, there is a distinct fixed point of the correctly trained deep network for every distinct output.

Next, we consider the flow in coupling-constant space close to a fixed point and, in particular, the stability properties of the fixed point as the number of layers is increased. We assume that the deeper the network the more accurately it is able to learn the classification problem. Close to the fixed point $\{g^*\}$, the relationship between the couplings in adjacent layers is, to leading order,
\begin{eqnarray}
g^{(k+1)}_\alpha(\{g^{(k)}\}) &=& g^{(k+1)}_\alpha(\{g^*\}) \nonumber\\
&& + \sum_\beta{(g^{(k)}_\beta - g^*_\beta)\left.\frac{\partial g^{(k+1)}_\alpha}{\partial g^{(k)}_\beta}\right|_{\{g^*\}}}.
\end{eqnarray}
Since $g^{(k+1)}_\alpha(\{g^*\}) = g^*_\alpha$, this becomes
\begin{equation}
g^{(k+1)}_\alpha - g^*_\alpha = \sum_\beta{T^*_{\alpha\beta}(g^{(k)}_\beta - g^*_\beta)},
\label{eq:coupling flow}
\end{equation}
where
\begin{equation}
T^*_{\alpha\beta} = \left.\frac{\partial g^{(k+1)}_\alpha}{\partial g^{(k)}_\beta}\right|_{\{g^*\}}
\label{eq:stability matrix}
\end{equation}
is the stability matrix of the fixed point. The stability properties of a fixed point determine whether it becomes sensitive to small changes in the input data as the number of layers is increased.

Define the eigenvalues and eigenvectors of the stability matrix by
\begin{equation}
\sum_\beta{T^*_{\alpha\beta} \phi^{(\mu)}_\beta} = \Lambda_\mu \phi^{(\mu)}_\alpha
\label{eq:stability eigenvalues}
\end{equation}
and express the couplings, $g_\alpha$, and the corresponding operators, $O_\alpha({\bf h})$, in the basis of eigenvectors of the stability matrix as
\begin{eqnarray}
g_\alpha &=& \sum_\mu{\phi^{(\mu)}_\alpha} \tilde{g}_\mu,\\
\tilde{O}_\mu({\bf h}) &=& \sum_\alpha{\phi^{(\mu)}_\alpha O_\alpha({\bf h})}.
\label{eq:operators and couplings in eigenbasis}
\end{eqnarray}
Then the effective Hamiltonian in this basis is
\begin{equation}
H^{(k)}_{\bf x} = \sum_\mu{\tilde{g}^{(k)}_\mu \tilde{O}_\mu({\bf h})}
\label{eq:Hamilton in eigenbasis}
\end{equation}
and, close to the fixed point, from eqs~(\ref{eq:coupling flow}) and (\ref{eq:stability eigenvalues}),
\begin{eqnarray}
H^{(k+1)}_{\bf x} - H^*_{\bf x} &=& \sum_\mu{(\tilde{g}^{(k+1)}_\mu - \tilde{g}^*_\mu) \tilde{O}_\mu ({\bf h})} \nonumber\\
                                &=& \sum_\mu{\Lambda_\mu (\tilde{g}^{(k)}_\mu - \tilde{g}^*_\mu) \tilde{O}_\mu ({\bf h})}.
\label{eq:Hamiltonian flow}
\end{eqnarray}
In the language of the RG, if $\Lambda_\mu>1$, $\tilde{O}_\mu({\bf h})$ is relevant and, if $\Lambda_\mu<1$, it is irrelevant. In the context of deep learning, the existence of a relevant operator at a fixed point creates a sensitivity to input data which generate a non-zero coupling to this operator. Such data will lead to a probability distribution for the output which increasingly deviates from the fixed-point distribution as the network is made deeper (more layers are added). This is the source of vulnerability of deep networks to tiny changes in the input data which excite a relevant operator, leading to the sequence of effective Hamiltonians converging to the wrong fixed point, i.e., misclassifying the data.

As it stands, this provides an explanation of the potential vulnerability of deep networks, but doesn't help to diagnose whether a particular network is susceptible. This is because, in practice, we only have access to the RBM weights, e.g., eq.~(\ref{eq:RBM}), which enable us to sample the hidden nodes and output nodes for a given input, but don't enable us to construct the sequence of effective Hamiltonians, or the fixed-point Hamiltonian. In the next section, we will explain how Monte Carlo RG methods may be used to estimate the stability matrix $T^*_{\alpha\beta}$, defined in eq.~(\ref{eq:stability matrix}), and how its eigenvalues behave in deep networks close to the fixed point. This will be sufficient to determine whether a particular trained deep network has any fixed points with relevant operators. 

It is worth noting that the existence of multiple fixed points (associated with a non-trivial classification problem) does not imply the existence of relevant operators at any of the fixed points. In the robust situation, where no fixed point has a relevant direction, data close to the boundary between two universality classes are simply ambiguous and do not represent a source of adversarial attack (because a human would be equally likely to misclassify them). 

\section{\label{sec:exponents} Stability of Deep Learning}

Having trained a deep network of RBMs on a classification problem whose exact solution is $p({\bf y}|{\bf x})$, we have available a sequence of RBMs given, for example, by eq.~(\ref{eq:RBM}), which enables us to use the conditional probability distributions in eq.~(\ref{eq:deepRBMprob}) to sample the hidden nodes in each layer and the output layer. Using this capability, we apply methods developed for Monte Carlo RG~\cite{Ma1976, Swendsen1979} to estimate the stability matrix for each fixed point in the space of effective Hamiltonians.

Denote by $\langle O_\gamma\rangle^{(k)}$ the expectation value of the operator $O_\gamma$ with respect to the effective Hamiltonian in layer $k$, eq.~(\ref{eq:Hamiltonian}), i.e.,
\begin{eqnarray}
\langle O_\gamma\rangle^{(k)} &=& {\rm Tr}_{\bf h} O_\gamma({\bf h}) q_k({\bf h}|{\bf x}) \nonumber\\
                              &=& \frac{{\rm Tr}_{\bf h} O_\gamma({\bf h}){\rm e}^{-H^{(k)}_{\bf x}({\bf h})}}{Z^{(k)}_{\bf x}}.
\label{eq:expectation value}
\end{eqnarray}
We start from the chain rule applied to $\langle O_\gamma\rangle^{(k+1)}$ close to a fixed point (i.e., $k\approx N$ for $N$ large),
\begin{eqnarray}
\frac{\partial \langle O_\gamma\rangle^{(k+1)}}{\partial g^{(k)}_\beta}  &=& \sum_\alpha{\frac{\partial g^{(k+1)}_\alpha}{\partial g^{(k)}_\beta}\frac{\partial \langle O_\gamma\rangle^{(k+1)}}{\partial g^{(k+1)}_\alpha}} \nonumber\\
&=& \sum_\alpha{T^{(k+1)}_{\alpha\beta}\frac{\partial \langle O_\gamma\rangle^{(k+1)}}{\partial g^{(k+1)}_\alpha} }.
\label{eq:chain rule}
\end{eqnarray}
Here $T^{(k+1)}_{\alpha\beta}$ is the approximation to the stability matrix $T^*_{\alpha\beta}$ in eq.~(\ref{eq:stability matrix}) obtained from the $k^{\rm th}$ layer, which should converge to $T^*_{\alpha\beta}$ for large $k$. Differentiating eq.~(\ref{eq:expectation value}) and using eq.~(\ref{eq:couplings}), we obtain for the term on the RHS of eq.~(\ref{eq:chain rule})
\begin{equation}
\frac{\partial \langle O_\gamma\rangle^{(k+1)}}{\partial g^{(k+1)}_\alpha} = \langle O_\gamma\rangle^{(k+1)}\langle O_\alpha\rangle^{(k+1)} - \langle O_\gamma O_\alpha\rangle^{(k+1)}.
\label{eq:in-layer correlations}
\end{equation}
To obtain a similar expression for the term on the LHS of eq.~(\ref{eq:chain rule}), we need to use $t_{k+1}$, the RBM for layer $k+1$, to relate $H^{(k+1)}_{\bf x}$ to $H^{(k)}_{\bf x}$, as in eq.~(\ref{eq:RGT}). In Monte Carlo RG, this is where the RG blocking step enters. We obtain
\begin{eqnarray}
\frac{\partial \langle O_\gamma\rangle^{(k+1)}}{\partial g^{(k)}_\beta} &=& \langle O_\gamma\rangle^{(k+1)}\langle O_\beta\rangle^{(k)} \nonumber\\
&& - {\rm Tr}_{{\bf hh}'} O_\gamma({\bf h}) t_{k+1}({\bf h}|{\bf h}') O_\beta({\bf h}') q_k({\bf h}'|{\bf x}) \nonumber\\
&=& \langle O_\gamma\rangle^{(k+1)}\langle O_\beta\rangle^{(k)} - \langle O_\gamma t_{k+1} O_\beta\rangle^{(k)}.\nonumber\\
&&
\label{eq:between-layer correlations}
\end{eqnarray}
Thus, with a suitable choice of operators, given a specific input ${\bf x}$, by sampling the hidden nodes in the upper layers of the network (large $k$) and computing a set of expectation values of these operators, we can construct a set of linear equations for the elements of the stability matrix, $T^{(k+1)}_{\alpha\beta} \approx T^*_{\alpha\beta}$,
\begin{eqnarray}
\sum_\alpha&&{T^{(k+1)}_{\alpha\beta}\left[\langle O_\gamma\rangle^{(k+1)}\langle O_\alpha\rangle^{(k+1)} - \langle O_\gamma O_\alpha\rangle^{(k+1)}\right]} \nonumber\\
&=&\langle O_\gamma\rangle^{(k+1)}\langle O_\beta\rangle^{(k)} - \langle O_\gamma t_{k+1} O_\beta\rangle^{(k)}.
\label{eq:linear eq}
\end{eqnarray}
Hence, we may obtain a set of successive approximations for the eigenvalues $\Lambda_\mu$ and eigenvectors $\phi^{(\mu)}_\alpha$ in eq.~(\ref{eq:stability eigenvalues}).

By varying the subset of operators used in eq.~(\ref{eq:linear eq}), we can test the convergence of the estimates of the largest eigenvalues of the stability matrix associated with a given layer of the network and, by computing the eigenvalues from successive layers, we can investigate whether increasing depth of the network is associated with scaling behaviour of the corresponding operators, as in eq.~(\ref{eq:Hamiltonian flow}). 

If this scaling behaviour is due to an eigenvalue of magnitude greater than one, then even a tiny component of the initial data, which creates a non-zero coupling to the corresponding eigenvector, will cause instability of the fixed point if the network is deep. In the next section, we will explain how to identify the specific perturbation of the data which does this.

\section{\label{sec:data} Generation of Adversarial Data}

We consider the exact conditional probability distribution, $p({\bf y}|{\bf x})$, and hence the trained network, $q_N({\bf y}|{\bf x})$, to be parametrised by the input data, ${\bf x}$, via the fixed-point Hamiltonian in eqs~(\ref{eq:fixed point Hamiltonian}) and (\ref{eq:fixed point}). The Fisher Information Matrix (FIM)~\cite{MachtaChachraTranstrumSethna2013, RajuMachtaSethna2017} is a metric which measures how a probability distribution changes along different directions in parameter space. If the fixed-point Hamiltonian has an unstable direction due to a relevant operator, then this will correspond to the direction in parameter space along which the probability distribution changes fastest, i.e., the stiffest direction. Hence, to generate adversarial data we need to compute the eigenvector associated with the largest eigenvalue of the Fisher information matrix for the fixed-point probability distribution. A tiny admixture of this component in the data will lead to erroneous classification if the network is too deep.

The FIM is defined in terms of the Kullback-Liebler divergence, $D_{\rm KL}$, which is itself a measure of how distinguishable two probability distributions $p_1$ and $p_2$ are, using data sampled from $p_1$:
\begin{equation}
D_{\rm KL}(p_1||p_2) = {\rm Tr}_{\bf x} p_1({\bf x}) \log \left[ \frac{p_1({\bf x})}{p_2({\bf x})} \right].
\label{eq:KL}
\end{equation}
This enables us to measure the dependence of $q_N({\bf y}|{\bf x})$ on ${\bf x}$ by considering two nearby data points, ${\bf x}$ and ${\bf x}'$, and expanding the Kullback-Liebler divergence as a Taylor series in their difference~\cite{MachtaChachraTranstrumSethna2013},
\begin{eqnarray}
D_{\rm KL}\left( q_N({\bf y}|{\bf x})||q_N({\bf y}|{\bf x}') \right) &=& {\rm Tr}_{\bf y} q_N({\bf y}|{\bf x}) \log \left[ \frac{q_N({\bf y}|{\bf x})}{q_N({\bf y}|{\bf x}')} \right] \nonumber\\
&=& \frac{1}{2} \sum_{ij}{(x'_i - x_i) F^{\scriptscriptstyle (N)}_{ij} (x'_j - x_j)} \nonumber\\
&& + \ldots,
\label{eq:KL expansion}
\end{eqnarray}
where, using eqs (\ref{eq:Hamiltonian}), (\ref{eq:couplings}) and (\ref{eq:expectation value}),
\begin{eqnarray}
F^{\scriptscriptstyle (N)}_{ij} &=& \left. \frac{\partial^2}{\partial x'_i \partial x'_j} D_{\rm KL}\left( q_N({\bf y}|{\bf x})||q_N({\bf y}|{\bf x}') \right) \right|_{{\bf x}' = {\bf x}} \nonumber\\
&=& \left\langle \frac{\partial H^{\scriptscriptstyle (N)}_{\bf x}}{\partial x_i} \frac{\partial H^{\scriptscriptstyle (N)}_{\bf x}}{\partial x_j} \right\rangle^{\scriptscriptstyle (N)}
 - \left\langle \frac{\partial H^{\scriptscriptstyle (N)}_{\bf x}}{\partial x_i} \right\rangle^{\scriptscriptstyle (N)} \left\langle \frac{\partial H^{\scriptscriptstyle (N)}_{\bf x}}{\partial x_j} \right\rangle^{\scriptscriptstyle (N)} \nonumber\\
&=& \sum_{\alpha\alpha'}{\frac{\partial g^{\scriptscriptstyle (N)}_\alpha}{\partial x_i} \frac{\partial g^{\scriptscriptstyle (N)}_{\alpha'}}{\partial x_j} \left[ \langle O_\alpha O_{\alpha'}\rangle^{\scriptscriptstyle (N)} - \langle O_\alpha \rangle^{\scriptscriptstyle (N)} \langle O_{\alpha'}\rangle^{\scriptscriptstyle (N)} \right]} \nonumber\\
&&
\label{eq:FIM}
\end{eqnarray}
is the FIM.

Having chosen a subspace of operators, $\{O_\alpha\}$, in which to express the effective Hamiltonians in each layer of the network (via eq.~(\ref{eq:couplings})), such that sufficiently precise estimates of the largest eigenvalues of the stability matrix associated with each layer, $T^{(k+1)}_{\alpha\beta}$, are obtained by the method in Section~\ref{sec:exponents}, we can compute the FIM using the chain rule:
\begin{equation}
\frac{\partial g^{(N)}_\alpha}{\partial x_i} = \sum_\beta{\left[ T^{(N)} \ldots T^{(2)} \right]_{\alpha\beta} \frac{\partial g^{(1)}_\beta}{\partial x_i}}.
\label{eq:product of Ts}
\end{equation}
The last derivative may be computed from eq.~(\ref{eq:first layer}) and the explicit form of the RBM for the first layer, e.g., eq.~(\ref{eq:RBM}), by expressing ${\bf h}_1^{\rm T} {\bf W}_1 {\bf x} + {\bf a}_1^{\rm T} {\bf h}$ in terms of $\{O_\alpha\}$ to determine $g^{(1)}_\beta ({\bf x})$. Alternatively, we may solve the system of linear equations in eq.~(\ref{eq:linear eq}) for the first layer:
\begin{eqnarray}
\sum_\alpha&&{\frac{\partial g^{(1)}_\alpha}{\partial x_i}\left[\langle O_\gamma\rangle^{(1)}\langle O_\alpha\rangle^{(1)} - \langle O_\gamma O_\alpha\rangle^{(1)}\right]} \nonumber\\
&=& {\rm Tr}_{\bf h} O_\gamma({\bf h}) \frac{\partial}{\partial x_i} t_1({\bf h}|{\bf x}).
\label{eq:linear eq for first layer}
\end{eqnarray}
Thus, computing the FIM involves translating the input data into the chosen coupling-constant subspace via the RBM in the first layer, taking the product of the stability matrices associated with each layer in eq.~(\ref{eq:product of Ts}), and computing a set of expectation values of operators in this subspace using the full network. The fact that the FIM is built from the product of stability matrices connects the scaling behaviour in eqs (\ref{eq:coupling flow}) and (\ref{eq:Hamiltonian flow}) with the sloppiness of the FIM spectrum observed in~\cite{MachtaChachraTranstrumSethna2013}, i.e., all but a fixed number of eigenvalues of the FIM are scaled to very small values, resulting in a fixed number of stiff (relevant) directions and many sloppy (irrelevant) directions. 

We may then obtain the eigenvector corresponding to the largest eigenvalue of the FIM. For deep networks which have an unstable fixed point, small perturbations of the corresponding input data in the direction of this eigenvector of the FIM are likely to result in the data being misclassified.

\section{\label{sec:conclusions} Conclusions}

In this paper we have extended the formal analogy between deep learning and the RG. This enabled us to interpret the classification problem in terms of a sequence of effective Hamiltonians associated with the layers of the network, whose couplings are determined by the input data, and the convergence of this sequence to a distinct fixed point for each distinct output of the classification problem. Input data with the same classification should all result in a flow in the coupling-constant space of these effective Hamiltonians which converges to the same fixed point. 

This exposed the possibility that a fixed point might have an unstable direction, associated with a relevant operator in the language of the RG. Small perturbations of the input data that create a non-zero overlap with such a relevant operator tend to be amplified exponentially by successive layers of the network, so that the flow diverges from the correct fixed point and the data may be misclassified. Fixed points with no relevant operators are not prone to this sort of error and will only misclassify data that is ambiguous even to a human.

Convergence of the sequence of effective Hamiltonians to a fixed point for a given subset of the input data is guaranteed by the training (assuming this converges to a good approximation to the exact conditional probability distribution). The existence of a relevant direction at the fixed point is associated with sensitivity to perturbations which take the data outside of this subset. The size of the amplification factor for the perturbation is greater the greater the depth of the network. Hence, the vulnerability to misclassification is directly due to the depth of a network.

The key result of this paper is a method for diagnosing whether a given trained network is vulnerable to adversarial attack, or simply misclassification due to noise in the data, based only on computing expectation values of operators in the hidden and output layers. The computations are fairly demanding, but may be justified for a network that is intended for use in safety-critical applications.

\end{document}